\documentclass[lettersize,journal]{IEEEtran}
\usepackage{amsmath,amsfonts}
\usepackage{array}
\usepackage[caption=false,font=normalsize,labelfont=sf,textfont=sf]{subfig}
\usepackage{textcomp}
\usepackage{stfloats}
\usepackage{url}
\usepackage{verbatim}
\usepackage{graphicx}
\usepackage{cite}

\usepackage{picinpar}
\usepackage{flushend}
\usepackage{colortbl}
\usepackage{soul}
\usepackage{multirow}
\usepackage{pifont}
\usepackage{color}
\usepackage{alltt}
\usepackage{enumerate}
\usepackage{siunitx}
\usepackage{breakurl}
\usepackage{epstopdf}
\usepackage{pbox}
\usepackage{float}
\usepackage{subfig}
\usepackage[colorlinks=true, urlcolor=magenta]{hyperref}
\usepackage{booktabs}
\usepackage{tikz}
\usepackage{lineno} 
\usepackage[ruled,vlined,linesnumbered]{algorithm2e}

\begin{document}

\title{ParaMaP: Parallel Mapping and Collision-free Motion Planning for Reactive Robot Manipulation}

\author{Xuewei Zhang, Bailing Tian, Kai Zheng, Yulin Hui, Junjie Lu and Zhiyu Li
  \thanks{This work was supported in part by the National Key R\&D Program of China under Grant 2023YFB4704900 and in part by the National Natural Science Foundation of China under Grant 62273249, Grant 62503354, Grant 62373268, and Grant 62373273, and in part by the China Postdoctoral Science Foundation under Grant Number 2024M762355. (Corresponding author: Zhiyu Li)}
  \thanks{Xuewei Zhang, Bailing Tian, Kai Zheng, Yulin Hui, Junjie Lu and Zhiyu Li are with the School of Electrical and Information
  Engineering, Tianjin University, Tianjin 300072, China (e-mail:
  zhangxuewei@tju.edu.cn; bailing\_tian@tju.edu.cn; zk545846621@tju.edu.cn; huiyulin@tju.edu.cn; lqzx1998@tju.edu.cn; lizhiyu@tju.edu.cn).}
}





\maketitle

\begin{abstract}
  Real-time and collision-free motion planning remains challenging for robotic manipulation in unknown environments due to continuous perception updates and the need for frequent online replanning. To address these challenges, we propose a parallel mapping and motion planning framework that tightly integrates Euclidean Distance Transform (EDT)-based environment representation with a sampling-based model predictive control (SMPC) planner. On the mapping side, a dense distance-field-based representation is constructed using a GPU-based EDT and augmented with a robot-masked update mechanism to prevent false self-collision detections during online perception. On the planning side, motion generation is formulated as a stochastic optimization problem with a unified objective function and efficiently solved by evaluating large batches of candidate rollouts in parallel within a SMPC framework, in which a geometrically consistent pose tracking metric defined on $\mathrm{SE}(3)$ is incorporated to ensure fast and accurate convergence to the target pose. The entire mapping and planning pipeline is implemented on the GPU to support high-frequency replanning. The effectiveness of the proposed framework is validated through extensive simulations and real-world experiments on a 7-DoF robotic manipulator. More details are available at: \href{https://zxw610.github.io/ParaMaP}{https://zxw610.github.io/ParaMaP}. 
\end{abstract}

\begin{IEEEkeywords}
  Perception and mapping, Motion planning, Sampling-based MPC, Robot manipulation, Parallel computing.
\end{IEEEkeywords}

\section{Introduction}
\IEEEPARstart{I}{n} recent years, robotic manipulation has been increasingly deployed in various industrial and service applications such as 3C assembly, point-of-use (POU) material handling and household assistance \cite{zhao2024survey}. These systems have demonstrated high precision and reliability in structured and well-modeled environments through modular programming and pre-defined task pipelines. With the rapid development of high-precision sensors such as depth cameras and 3D LiDAR, as well as advances in perception and motion planning, robots are expected to autonomously perform interactive manipulation and human-robot collaboration tasks in dynamic and unstructured environments \cite{liu2024integrating}. Therefore, it is essential to efficiently represent environment, while designing real-time motion planners to ensure safe and reactive operation. 

Voxel-based environment representation methods are well established \cite{GridMap}. They estimate occupancy probabilities over the volume of 3D space, enabling explicit representation of occupied, free, and unknown regions. For a more refined spatial representation, the Truncated Signed Distance Function (TSDF) \cite{Kinectfusion} and the Euclidean Signed Distance Function (ESDF) \cite{Voxblox, Fiesta} have been successively developed to encode signed distance values, enabling more accurate and continuous geometric modeling. TSDF and ESDF are widely adopted in gradient-based trajectory optimization because they provide continuous distance fields to obstacles, which allow collision costs to be formulated as differentiable terms in the objective function. However, constructing such a distance field is computationally expensive and has become the main bottleneck preventing these methods from being effectively deployed on resource-limited platforms. To improve computational efficiency, several studies \cite{nvblox, GIE-mapping} have exploited GPU-based parallel computing to accelerate the construction of environment maps.

In addition to environment representation, motion planning plays a central role in achieving real-time and collision-free robot motion. The problem of robot motion planning has been actively investigated. The Open Motion Planning Library (OMPL) \cite{OMPL} is one of the most widely used open-source frameworks for this purpose. It offers a broad set of sampling-based path planning algorithms such as RRTConnect, RRT*, and PRM, together with standardized interfaces for integrating planners into robotic systems. While such path planners can efficiently find feasible paths, they are unable to generate smooth trajectories suitable for execution on high-order dynamical systems. To address this limitation, gradient-based optimization methods \cite{CHOMP, Ego-planner, xu2024fixed} have been developed, which formulate motion planning as a continuous optimization problem to generate smooth and dynamically feasible trajectories. Nevertheless, when the objective function involves discontinuities, explicit gradients may not be available. In such cases, gradient-free optimization approaches, exemplified by sampling-based model predictive control (SMPC)\cite{STOMP, Storm}, have been proposed, leveraging stochastic trajectory sampling and parallel computation to achieve robust real-time performance. However, these algorithms have been primarily validated in structured environments where obstacle geometries are known or signed distance fields are pre-computed, limiting their applicability in unknown and dynamically changing environments.

Motivated by the above analysis, we present \textbf{ParaMaP}, a \textbf{Para}llel \textbf{Ma}pping and \textbf{P}lanning framework for reactive robot manipulation in environments with unknown obstacles. Unlike traditional ray-casting-based mapping approaches \cite{Voxblox, Fiesta}, ParaMaP constructs an occupancy grid using a parallel voxel projection strategy and converts it into a Euclidean Distance Transform (EDT) via a Felzenszwalb--Huttenlocher (FH) squared distance transform. Building on the resulting distance-field representation, motion planning is formulated as a stochastic optimization problem with a unified cost function that integrates collision avoidance, smoothness, dynamic feasibility, and a geometrically consistent pose error defined on $\mathrm{SE}(3)$. The entire mapping and planning pipeline is implemented on the GPU using CUDA to support high-frequency replanning. The main contributions of this work are summarized as follows:

\begin{enumerate}
  \item We propose a novel sampling-based motion planning formulation for robotic manipulation that integrates distance-field-based collision cost with a Lie-algebra-based pose tracking objective on $\mathrm{SE}(3)$ within a stochastic MPC framework, enabling safe and reactive motion generation in environments with unknown obstacles.
  
  \item We introduce a Lie-algebra-based pose error metric on $\mathrm{SE}(3)$ and incorporate it into the proposed SMPC formulation, providing a geometrically consistent treatment of rotational and translational deviations and enabling fast convergence to target poses subject to practical execution constraints.
  
  \item We develop an efficient parallel mapping and Euclidean Distance Field construction method based on a gather-then-transform strategy and robot-masked updates, which supports low-latency distance queries and reliable collision avoidance during online replanning.
  
  \item The proposed mapping and planning method achieves real-time and high-frequency replanning for reactive robot manipulation by leveraging a unified CUDA-based architecture that exploits GPU parallelism.
\end{enumerate}

\section{Related work}
\subsection{Environment Mapping for planning}
Environment mapping is a fundamental prerequisite for robot motion planning, as it provides a geometric representation of the workspace necessary for collision avoidance and trajectory optimization. The grid map \cite{OctoMap} is a common representation that discretizes the environment into regular cells, where each cell stores an occupancy probability indicating whether it is free, occupied, or unknown. This representation is simple and widely used for collision checking in hard-constrained robot trajectory generation \cite{Teach-repeat-replan}. In \cite{Skeleton}, a sparse topological skeleton graph was proposed to compactly encode the free-space structure, but it relies on a fully known environment map. Besides, \cite{GP-Frontier} employs a variational sparse Gaussian Process to infer a local occupancy surface from onboard LiDAR measurements and selects frontier sub-goals based on model uncertainty, enabling lightweight local navigation through an implicit environmental representation. 

The aforementioned approaches do not provide dense and continuous distance information to obstacles, which is essential for optimization-based planning \cite{DEMO-PAST}. To address this limitation, the TSDF \cite{Chisel} was introduced to store the signed projective distance along sensor rays to the observed surface. Compared with the TSDF, the ESDF computes true Euclidean distances to the nearest obstacles over the mapping region, making it more suitable for motion planning \cite{CHOMP}. Nevertheless, maintaining an accurate and up-to-date ESDF for reactive motion planning remains computationally demanding, particularly in dynamic or partially unknown environments. To ensure real-time performance, \cite{Ewok} maintains a fixed-size local ESDF map that slides with the robot. Additionally, the approaches in \cite{Voxblox}  and \cite{Fiesta}  build an incremental EDT online by leveraging elaborated data structures and update strategy. With the advancement of CUDA, some works \cite{nvblox, GIE-mapping} have employed parallel acceleration to construct and maintain EDT maps, aiming to bridge the gap between efficiency and accuracy.

\subsection{Robot Reactive Motion Planning}
Robot reactive motion planning is crucial for safety and is typically formulated as an optimization problem, which can be further classified into two categories depending on whether the optimization process relies on gradient information. Gradient-based motion planning methods are well-explored. Schulman et al. \cite{schulman2013finding} used sequential convex optimization with continuous-time safety constraints to find locally optimal, collision-free trajectories. Several studies \cite{Teach-repeat-replan, Cmpcc} identify obstacle-free corridors as a sequence of convex shapes and then generate safe trajectories within these free spaces using convex optimization such as Second-Order Cone Program (SOCP) and quadratic program (QP). In \cite{Ego-planner, Minco, SUPER}, motion planning is formulated as a
nonlinear unconstrained optimization problem with a unified value function that includes penalty terms for collision, control, and constraint violations.

Although gradient-based methods can be implemented with low memory consumption and solved using off-the-shelf solvers such as NLopt\footnote{\url{https://nlopt.readthedocs.io}}, explicit Jacobian matrices may not exist when the value function contains discontinuous components. To address this, sampling-based optimization methods, especially SMPC, that do not rely on gradient information have been widely developed in recent research. Kalakrishnan et al. \cite{STOMP} approached trajectory planning as a stochastic optimization problem, using path integrals to derive a stochastic gradient. They then solved the problem by weighting noisy trajectories and combining them to produce an updated trajectory with a lower cost. To improve the computational efficiency of SMPC, \cite{Storm, MPPI} introduced GPU parallelization, enabling real-time planning for high-dimensional robotic manipulation tasks. Subsequently, some works based on SMPC, such as robust SMPC \cite{RMPPI} and risk-aware SMPC \cite{RAPA-Planner}, were proposed to further enhance robustness.

\section{Parallel Occupancy Grid Map and Euclidean Distance Field Construction}
In this section, we present a parallel occupancy grid map and EDT construction method that enables efficient GPU execution. The proposed approach overcomes the inherent parallelization challenges of conventional ray-casting-based mapping pipelines by employing a voxel projection strategy to construct the occupancy grid, followed by a gather-then-transform EDT computation from the resulting grid. As a result, low-latency map updates can be achieved to support high-frequency motion replanning.

\subsection{Occupancy Grid Map Construction}
The most common method for constructing an occupancy grid map (OGM) is ray casting \cite{Ewok, Fiesta}, where rays are emitted from the sensor's position toward each measured point. For cameras, the measured points are obtained by projecting each pixel into 3D space using the depth value and intrinsic parameters, while LiDAR directly provides 3D spatial points, i.e., point clouds.
The intersection of each ray with objects in 3D space is used to update the occupancy status of the corresponding grid cells. The computational complexity of the ray casting algorithm is $O\left( {Nd/r} \right)$, where $N$ is the number of measured surface points, $d$ is the average travel distance, and $r$ is the voxel size. In addition, ray casting is difficult to execute in parallel due to the possibility of multiple rays passing through the same grid cell and attempting to update the same memory location, leading to I/O conflicts and decreased computational efficiency.

\begin{figure}[!t]
  \centering
  \subfloat[]{\includegraphics[width=1.85in]{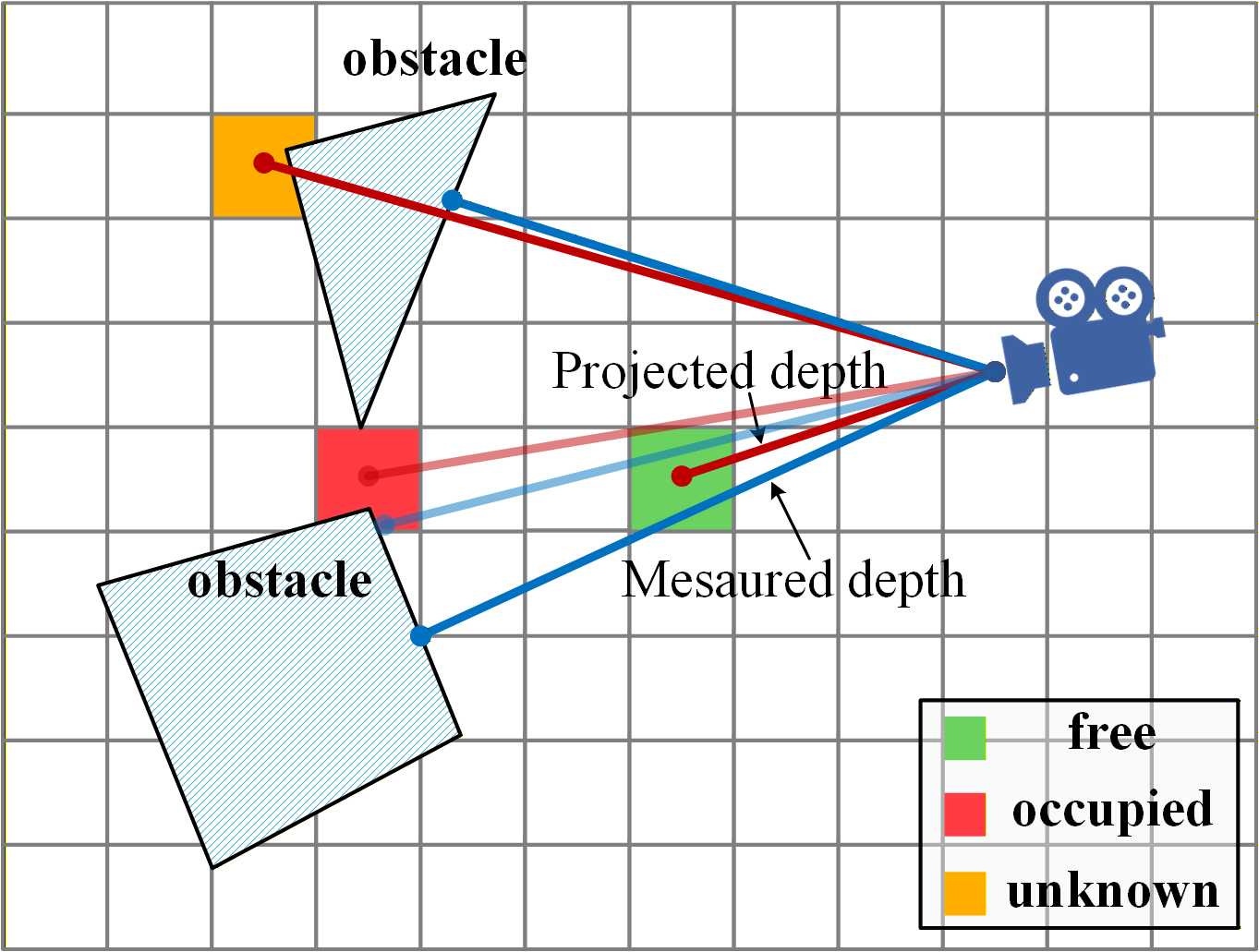}
  \label{fig_first_case}}
  \subfloat[]{\includegraphics[width=1.5in]{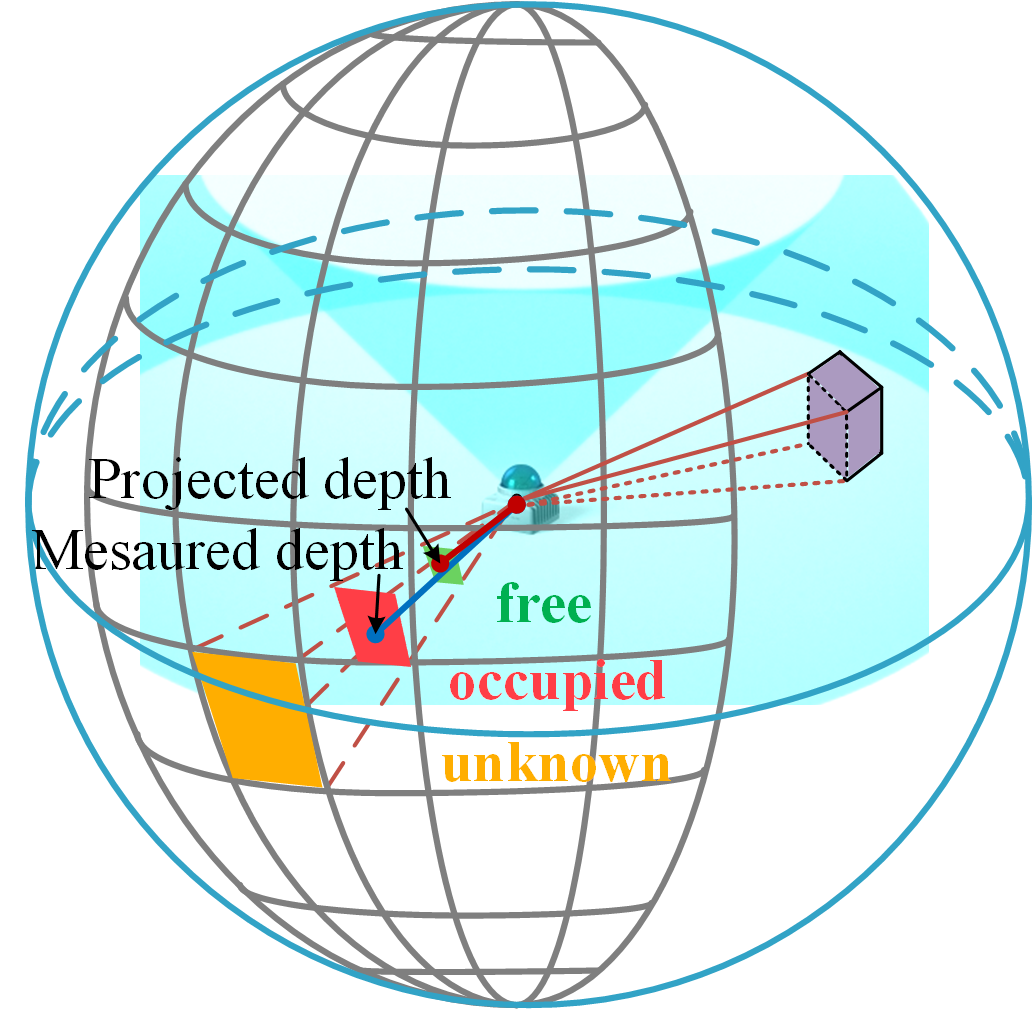}
  \label{fig_second_case}}
  \caption{Illustration of the voxel projection method for occupancy updating. (a) Camera-based projection of 3D voxels onto the image plane. (b) LiDAR-based projection of 3D voxels and point clouds onto spherical coordinates.} 
  \label{fig_voxel_projection}
\end{figure}

Inspired by the voxel projection strategy introduced in \texttt{GIE-mapping} \cite{GIE-mapping}, we adopt a voxel-centric occupancy update scheme that is more amenable to parallel execution on the GPU. Instead of tracing rays through the voxel grid, the proposed approach updates the occupancy state of each voxel by projecting its 3D volume onto the sensor measurement domain, i.e., the 2D image plane for cameras or spherical coordinates for LiDAR. The detailed projection process is illustrated in Fig.~\ref{fig_voxel_projection}, which shows how each voxel is associated with the corresponding sensor measurement. This voxel-centric formulation enables independent occupancy evaluation for each voxel, thereby avoiding write conflicts caused by multiple rays intersecting the same grid cell and improving parallel efficiency. Based on the projected measurements, each voxel is subsequently classified as occupied, free, or unknown according to the following update rules.

$\bullet$  Occupied: If the projected depth closely matches the measured depth within a predefined tolerance, the voxel is considered to lie on the observed surface. In this case, the voxel is marked as occupied, and its occupancy probability is increased.

$\bullet$ Free: If the projected depth is smaller than the measured depth, the voxel lies in front of the observed surface along the same projection direction. The voxel is therefore classified as free volume, and its occupancy probability is reduced.

$\bullet$ Unknown: If the projected depth exceeds the measured depth, the voxel falls behind the observed surface and is occluded by an obstacle. Such voxels cannot be reliably interpreted from the current measurement and remain unchanged and are treated as unknown.

In practice, our method allows each voxel to be assigned to a separate thread. Compared to ray casting, the computational complexity of voxel projection is $O\left( {n} \right)$, where $n$ denotes the number of updated voxels. With parallel execution, the overall runtime is substantially reduced.

\subsection{Euclidean Distance Field Construction}
After constructing the occupancy grid map, a continuous measure of obstacle proximity is required to support collision checking and safe motion planning. To obtain this, each occupied voxel is treated as a zero-distance source, whereas free or unknown voxels are initialized with a large value, forming the input to the Euclidean Distance Field (EDF). The EDF assigns to every voxel the squared distance to its nearest obstacle and provides a smooth geometric representation well suited for trajectory optimization. To compute the field efficiently, we employ the classical three-pass separable 3D EDT, in which the 1D Felzenszwalb--Huttenlocher (FH) squared distance transform is applied sequentially along the $y$-, $x$-, and $z$-axes. This separable formulation guarantees exactness and achieves an overall computational complexity of $O(n)$ for a map of $n = {N_x} \times {N_y} \times {N_z}$ voxels, where $N_x$, $N_y$, and $N_z$ denote the number of voxels along the three axes.

A key difference from \texttt{GIE-mapping} lies in how the 1D FH transforms along the $x$- and $z$-directions are performed. In \texttt{GIE-mapping}, the 3D voxel grid is explicitly permuted using the CUDA Tensor Transpose (cuTT) library\footnote{\url{https://github.com/ap-hynninen/cutt}} such that the data along each transform direction becomes contiguous in memory (e.g., $XYZ \to ZYX$ before the $x$-pass). This is necessary for the original implementation because each FH kernel assumes that its input line is stored in a single, contiguous region of memory. Nevertheless, these full-grid permutations incur significant global-memory traffic and become the dominant cost when the local update region is relatively small.
In the proposed approach, the FH transforms are executed directly on the native $XYZ$ memory layout without any tensor permutation. For the $x$-pass, each CUDA thread is assigned a fixed $(y,z)$ coordinate and process the 1D line

\begin{equation}
  g\left( {x,y,z} \right),x = 0,1, \cdots ,{N_x} - 1,
\end{equation}
which is stored in global memory with a stride of
\begin{equation}
  strid{e_x} = {N_y} \times {N_z}.
\end{equation}

Instead of transposing the entire tensor, each thread gathers this stride sequence from global memory into a compact per-thread shared-memory buffer

\begin{equation}
  \mathrm{buf}_x\left[ x \right] = g\left( {x,y,z} \right),x = 0,1, \cdots ,{N_x} - 1.
\end{equation}

The FH transform is then performed locally and sequentially inside this contiguous buffer. The same procedure is used in the $z$-pass. For a fixed $(x,y)$ coordinate, the thread fetches the line

\begin{equation}
  g\left( {x,y,z} \right),z = 0,1, \cdots ,{N_z} - 1,
\end{equation}
from global memory with a stride

\begin{equation} 
  strid{e_z} = {N_y},
\end{equation}
copies it into a contiguous shared-memory buffer, and performs the FH transform locally. Since all FH operations are performed on per-thread compact buffers, the transform remains mathematically identical to the standard FH algorithm while avoiding all global tensor permutations.

This gather-then-transform strategy eliminates the need for cuTT-based tensor transpositions and significantly reduces global-memory movement. Memory accesses during the gather stage are structured to maintain efficient, coalesced global-memory reads across threads, while shared-memory usage remains minimal and bounded by the 1D line length. In addition, rather than maintaining a globally integrated EDT, the distance field is recomputed only within a local volume at each update cycle, keeping the computational overhead lightweight. As a result, the proposed EDT module achieves substantially lower latency and provides a streamlined, highly responsive distance-field computation well suited for real-time reactive robot manipulation.

\section{Motion Planning with Sampling-based Model Predictive Control}

\subsection{Problem Formulation}
Consider the task of driving a robot from an initial joint configuration $\boldsymbol{q}_0$  to a desired end-effector (EE) pose $\boldsymbol{T}_g \in \mathrm{SE}\left( 3 \right)$. To achieve this, the planner needs to compute a feasible trajectory that respects the robot's kinematic limits, avoids collisions, and can be efficiently replanned in real time as new sensor information becomes available. In this work, we formulate the motion planning problem as a time discretized stochastic optimization problem.

Let $\boldsymbol{q}_k \in \mathbb{R}^n$ denote the robot's joint configuration at time step $k$, where $n$ is the number of actuated degrees of freedom (DoF) of the manipulator. The associated Cartesian pose, obtained through the forward kinematics, is given by
\begin{equation}
    \boldsymbol{T}_k = f_{\mathrm{FK}}(\boldsymbol{q}_k) =
    \begin{bmatrix}
      \boldsymbol{R}(k) & \boldsymbol{t}(k) \\
      \boldsymbol{0}^\top & 1
    \end{bmatrix}
    \in \mathrm{SE}(3) ,
    \label{eq:forward_kinematics}
\end{equation}
where $\boldsymbol{R}(k) \in \mathrm{SO}(3)$ and $\boldsymbol{t}(k) \in \mathbb{R}^3$. Furthermore, we denote by $\boldsymbol{T}^i_k \in \mathrm{SE}(3)$ the Cartesian pose of the $i$-th robot link at time step $k$, where $i = 0,\ldots,n$ for an $n$-DoF manipulator. The end-effector pose is given by $\boldsymbol{T}_k^{\mathrm{ee}}$, which corresponds to the last link pose in the forward kinematics chain; that is, $\boldsymbol{T}_k^{\mathrm{ee}} = \boldsymbol{T}^n_k$.

The robot dynamics over a finite prediction horizon $H$ are described by
\begin{equation}
  {{\boldsymbol{q}}_{k + 1}} = f({{\boldsymbol{q}}_k},{{\boldsymbol{u}}_k},\boldsymbol{\epsilon}_k),
  \label{eq:dynamic}
\end{equation}
where $\boldsymbol{u}_k$ represents the control input in joint space, taken here as the joint accelerations, and $\boldsymbol{\epsilon}_k$ represents process noise or model uncertainty. 

The objective is to solve a control sequence in joint
\begin{equation}
  \mathbf{U} = \{ \boldsymbol{u}_0, \boldsymbol{u}_1, \ldots, \boldsymbol{u}_{H-1} \}
\end{equation}
that minimizes the expected finite-horizon cost
\begin{equation}
  J(\mathbf{U}) = 
  \mathbb{E}_{\boldsymbol{\epsilon}_{0:H-1}} \left[
      \ell_f(\boldsymbol{T}_H, \boldsymbol{T}_g)
      + \sum_{k=0}^{H-1} \ell(\boldsymbol{q}_k, \boldsymbol{x}_k, \boldsymbol{u}_k)
  \right],
  \label{eq:cost}
\end{equation}
where $\ell_f(\boldsymbol{T}_H, \boldsymbol{T}_g)$ is the terminal cost that penalizes the deviation of the terminal end-effector pose from the goal, and $\ell(\boldsymbol{q}_k, \boldsymbol{x}_k, \boldsymbol{u}_k)$ is the running cost that accounts for intermediate objectives such as smoothness and collision avoidance. The expectation in \eqref{eq:cost} arises from the stochastic system evolution induced by the process noise $\boldsymbol{\epsilon}_k$, as well as from the sampling-based evaluation used to approximate the optimal control sequence.

The optimal control sequence is obtained by solving
\begin{equation}
  \label{eq:opt_problem}
  \begin{aligned}
    \mathbf{U}^\star
    = \;& \arg\min_{\mathbf{U}} \; J(\mathbf{U}) \\
    \text{s.t. } 
    & \boldsymbol{q}_{k+1} = f(\boldsymbol{q}_k, \boldsymbol{u}_k, \boldsymbol{\epsilon}_k), \\
    & \boldsymbol{T}_k = f_{\mathrm{FK}}(\boldsymbol{q}_k), \\
    & \boldsymbol{q}_k \in \mathcal{Q}_{\mathrm{env}}^{\mathrm{free}} 
        \cap \mathcal{Q}_{\mathrm{self}}^{\mathrm{free}}, \\
    & \boldsymbol{u}_k \in \mathcal{U},
    \qquad k = 0,\dots,H-1,
  \end{aligned}
\end{equation}
where $\mathcal{Q}_{\mathrm{env}}^{\mathrm{free}}$ denotes the set of robot configurations that are free of collisions with the environment, as determined by the EDF constructed in Sec.~III-B. Likewise, $\mathcal{Q}_{\mathrm{self}}^{\mathrm{free}}$ denotes the set of configurations that avoid self-collision among the robot's links. The admissible control set $\mathcal{U}$ specifies the limits on joint positions, velocities, and actuation.

\subsection{Cost Function Construction}
The overall objective defined in~\eqref{eq:cost} consists of a running cost accumulated along the prediction horizon and a terminal cost applied at the final step. The running cost is composed of five components—pose tracking, collision avoidance, joint limits, trajectory smoothness, and null-space regularization—each evaluated at every step $k = 0,\dots,H-1$ to promote smooth, safe, and dynamically feasible motion. In addition, a terminal cost is imposed at $k = H$ to ensure convergence of the end-effector to the desired goal pose. The detailed formulations of all cost components are provided in the following subsections.

\subsubsection{Pose Tracking Cost}
To steer the end-effector toward the desired pose $\boldsymbol{T}_g$, we adopt a geometrically consistent error metric defined directly on the Lie group $\mathrm{SE}(3)$. The relative transformation from the goal frame to the current end-effector frame is
\begin{equation}
  \Delta \boldsymbol{T}(k) = \boldsymbol{T}_g^{-1}\boldsymbol{T}_k^{\mathrm{ee}}
  =
  \begin{bmatrix}
    \boldsymbol{R}_\Delta(k) & \boldsymbol{t}_\Delta(k) \\
    \boldsymbol{0}^\top & 1
  \end{bmatrix},
\end{equation}
with $\boldsymbol{R}_\Delta(k)=(\boldsymbol{R}_g)^\top \boldsymbol{R}(k)$,  $\boldsymbol{t}_\Delta(k)=(\boldsymbol{R}_g)^\top(\boldsymbol{t}(k)-\boldsymbol{t}_g)$.

A unified $6$-D pose error is obtained by applying the matrix logarithm on $\Delta \boldsymbol{T}(k)$:
\begin{equation}
  \log(\Delta \boldsymbol{T}(k)) =
  \begin{bmatrix}
  [\boldsymbol{\omega}(k)]_\times & \boldsymbol{v}(k) \\
  \boldsymbol{0}^\top & 0
  \end{bmatrix}\in\mathfrak{se}(3),
\end{equation}
where $\boldsymbol{\omega}(k)\in\mathbb{R}^3$ is the rotation error vector satisfying $\boldsymbol{R}_\Delta(k)=\exp([\boldsymbol{\omega}(k)]_\times)$, and $\boldsymbol{v}(k)\in\mathbb{R}^3$ is the translational component of the Lie-algebraic error obtained through
\begin{equation}
  \boldsymbol{t}_\Delta(k)=\boldsymbol{V}(\theta(k))\,\boldsymbol{v}(k),  \theta(k)=\|\boldsymbol{\omega}(k)\|,
\end{equation}
with
\begin{equation}
  \boldsymbol{V}(\theta)=\boldsymbol{I}_3+\frac{1-\cos\theta}{\theta^2}[\boldsymbol{\omega}]_\times
  +\frac{\theta-\sin\theta}{\theta^3}[\boldsymbol{\omega}]_\times^2.
\end{equation}

The resulting SE(3) pose error vector is then
\begin{equation}
  \boldsymbol{\xi}(k)=\begin{bmatrix} \boldsymbol{v}(k) \\ \boldsymbol{\omega}(k) \end{bmatrix}\in\mathbb{R}^6.
\end{equation}

Since rotational and translational errors have different physical units and magnitudes, a positive definite weighting matrix $\boldsymbol{Q}_k$ is introduced to properly scale their respective contributions within the objective. The pose tracking cost is thus formulated as
\begin{equation}
  J_{\mathrm{pose}}
  =
  \sum_{k=0}^{H-1}
  \frac{1}{2}\,\boldsymbol{\xi}(k)^\top \boldsymbol{Q}_k\, \boldsymbol{\xi}(k).
\end{equation}

\noindent\textbf{Remark 1.} \textit{
  This Lie-algebra-based formulation provides a unified and geometrically meaningful measure of pose discrepancy on $\mathrm{SE}(3)$, ensuring consistent treatment of both rotational and translational deviations, which is not explicitly captured by the pose error formulation adopted in \texttt{STORM} \cite{Storm}.
}

\subsubsection{Collision Avoidance Cost}
\begin{figure}[!t]
  \centering
  \subfloat[]{\includegraphics[width=1.6in]{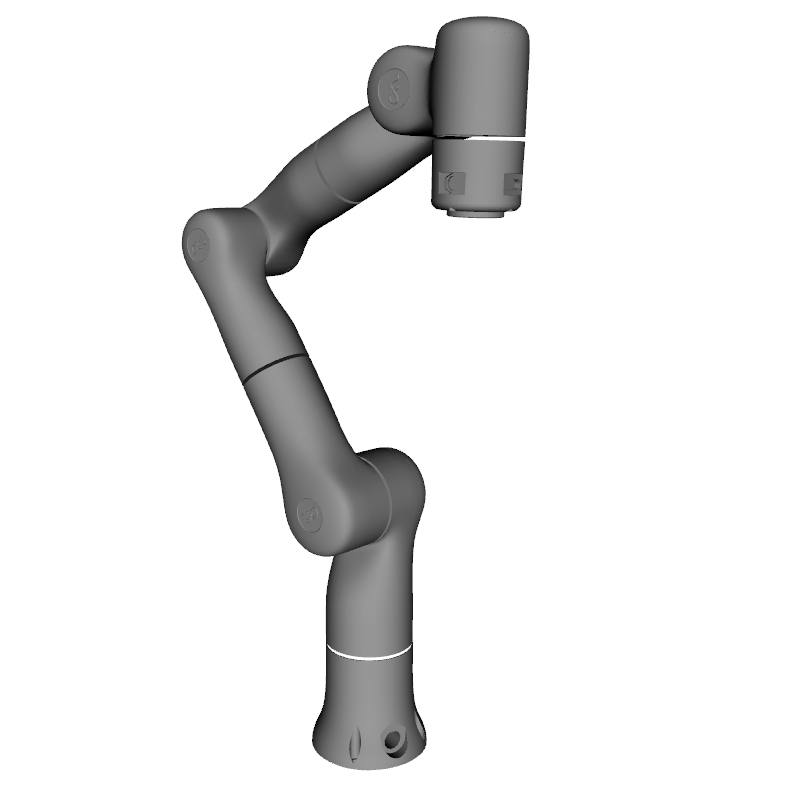}
  \label{fig_first_case}}
  \subfloat[]{\includegraphics[width=1.6in]{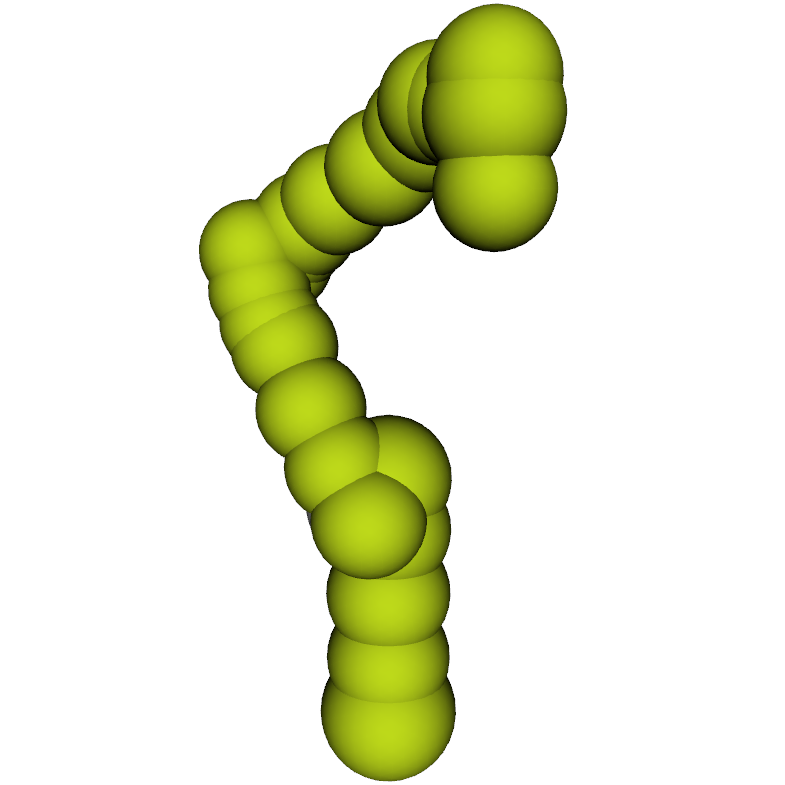}
  \label{fig_second_case}}
  \caption{Robot geometry representations. (a) The original mesh model of the robot rendered in RViz. (b) The simplified collision model constructed by approximating each link with a set of $n_s$ spheres.} 
  \label{fig_robot_spheres}
\end{figure}

Following \texttt{STORM} \cite{Storm}, we approximate the robot's geometry using a set of $n_s$ spheres, as illustrated in Fig.~\ref{fig_robot_spheres}. Let $\boldsymbol{p}_{i,k} \in \mathbb{R}^3$ denote the center of sphere $i$ at time step $k$, and let $r_i > 0$ be its radius. The workspace is represented by a ESDF $\phi:\mathbb{R}^3\!\rightarrow\!\mathbb{R}$, where $\phi(\boldsymbol{x})$ returns the signed distance from point $\boldsymbol{x}$ to the closest obstacle surface. The signed distance between sphere $i$ and the environment at time step $k$ is thus
\begin{equation}
    d^{\mathrm{env}}_{i,k}
    =
    \phi\!\left(\boldsymbol{p}_{i,k}\right) - r_i. 
\end{equation}

For self-collision, we consider a set $\mathcal{S}$ of sphere pairs $(i,j)$ that may come into contact under the robot's kinematics constraints. The signed distance between two spheres $(i,j) \in \mathcal{S}$ at time step $k$ is given by
\begin{equation}
    d^{\mathrm{self}}_{(i,j),k}
    =
    \left\|
        \boldsymbol{p}_{i,k} - \boldsymbol{p}_{j,k}
    \right\|
    -
    \left(r_i + r_j\right).
\end{equation}

To activate the collision penalty smoothly before contact, we introduce an activation distance $d_{\mathrm{act}}>0$. The corresponding violation terms for environment and self-collision distances are defined as
\begin{equation}
  \small
  \begin{alignedat}{1}
  \delta_{i,k}^{\mathrm{env}} & \;=\; \max\!\left(0,\, d_{\mathrm{act}} - d^{\mathrm{env}}_{i,k}\right), \\
  \delta_{(i,j),k}^{\mathrm{self}} & \;=\; \max\!\left(0,\, -d^{\mathrm{self}}_{(i,j),k}\right).
  \end{alignedat}
\end{equation}
  
The collision-avoidance cost accumulated over the prediction horizon is then
\begin{equation}
  \small
  \label{eq:collision_cost_final}
  J_{\mathrm{coll}}
  =
  \sum_{k=0}^{H-1}
  \left(
  \sum_{i=1}^{n_s}
      w_{\mathrm{env}}
      \big(\delta_{i,k}^{\mathrm{env}}\big)^2
  \;+\;
  \sum_{(i,j)\in\mathcal{S}}
      w_{\mathrm{self}}
      \big(\delta_{(i,j),k}^{\mathrm{self}}\big)^2
  \right),
\end{equation}
where $w_{\mathrm{env}}, w_{\mathrm{self}}>0$ are penalty weights for environment and self-collision avoidance.

In practice, depth sensors inevitably observe parts of the robot body, causing its geometry to be fused into the workspace EDF and mistakenly treated as an external obstacle during planning. A simple remedy is to subtract the robot model from the depth map or point cloud using its known kinematic skeleton. However, such geometry-level subtraction introduces substantial regions of missing depth, and subsequent filtering or inpainting attempts cannot reliably recover the true environment geometry.

\noindent\textbf{Remark 2.} \textit{
  Instead of operating at the raw depth or geometry reconstruction level, we prevent the robot geometry from being integrated into the occupancy map during sensor fusion. At each map update, all depth returns whose back-projected 3D points fall inside the robot's sphere model are discarded, and any voxels covered by the sphere model are explicitly reset to a non-occupied state in the occupancy layer. The EDF is then computed from this cleaned occupancy map, such that distances are defined only with respect to external obstacles, without requiring any depth reconstruction or surface completion.
  As a result, the distance field intentionally does not encode distance information relative to the robot surface, which differs from the assumptions commonly made in gradient-based trajectory optimization that depend on smooth and complete distance gradients. In contrast, the SMPC framework relies only on point-wise distance queries and does not require distance gradients, making it naturally compatible with the proposed robot-masked distance field formulation.
}

\subsubsection{Joint Limit Cost}
To ensure the planned trajectory stays within the robot's feasible
joint-space limits, we incorporate a joint limit cost that penalizes violations in joint position, velocity, and acceleration. Let 
$x_{i,k} \in \{q_{i,k}, \dot{q}_{i,k}, \ddot{q}_{i,k}\}$ denote the state of joint $i$ at time step $k$, and let $[x_i^{\min}, x_i^{\max}]$ represent its admissible range. In practice, this interval is slightly tightened to 
$[\,x_i^{\min,\varepsilon},\; x_i^{\max,\varepsilon}\,]$, where 
$x_i^{\min,\varepsilon}=x_i^{\min}+\varepsilon_i$ and 
$x_i^{\max,\varepsilon}=x_i^{\max}-\varepsilon_i$ with 
$\varepsilon_i>0$ a small activation margin applied proportionally to the 
original range. This tightening activates the penalty ahead of the hard limits, ensuring a smooth transition as the state approaches the boundary. The violation for joint $i$ is compactly expressed as
\begin{equation}
  \small
  \delta_i(x_k)
  =
  \max\!\left(0,\, x_{i,k} - x_i^{\max,\varepsilon}\right)
  +
  \min\!\left(0,\, x_{i,k} - x_i^{\min,\varepsilon}\right).
\end{equation}

The joint-limit cost accumulated over the entire prediction horizon is
\begin{equation}
  \label{eq:joint_limit_cost_final}
  J_{\mathrm{lim}}
  =
  \sum_{k=0}^{H-1}
  \sum_{i=1}^{n}
  \Big(
  w_q\,\delta_i(q_k)^2
  +
  w_{\dot{q}}\,\delta_i(\dot{q}_k)^2
  +
  w_{\ddot{q}}\,\delta_i(\ddot{q}_k)^2
  \Big),
\end{equation}
where $w_q$, $w_{\dot{q}}$, and $w_{\ddot{q}}$ are positive penalty weights.

\subsubsection{Smoothness Cost}
To promote smooth joint-space motion, we introduce a smoothness cost that 
penalizes excessive joint acceleration. Accordingly, the smoothness cost 
accumulated over the prediction horizon is defined as
\begin{equation}
  J_{\mathrm{sm}}
  =
  \sum_{k=0}^{H-1}
  \sum_{i=1}^{n}
  w_s\, \ddot{q}_{i,k}^{\,2},
  \label{eq:smoothness_acc_final}
\end{equation}
where $w_s > 0$ is the acceleration smoothing weight.

\subsubsection{Null-Space Regularization Cost}
To avoid kinematic singularities and reduce self-collision risk, we introduce a null-space regularization term that biases the robot toward a preferred reference configuration. Let $q^{\mathrm{ref}}_i$ denote the reference posture for joint $i$. The null-space regularization cost accumulated over the prediction horizon is
\begin{equation}
  J_{\mathrm{ns}}
  =
  \sum_{k=0}^{H-1}
  \sum_{i=1}^{n}
  w_{\mathrm{ns}}\,
  \big(q_{i,k} - q^{\mathrm{ref}}_i\big)^{2},
\end{equation}
where $w_{\mathrm{ns}}>0$ is the null-space weight.

It is worth noting that the end-effector pose $\boldsymbol{T}_g \in \mathrm{SE}(3)$ is $6$-D whereas the robot has 7-DoF in our configuration, multiple joint configurations can achieve the same target pose. Without regularization, the optimizer may drift toward near-singular or otherwise undesirable  configurations. The null-space regularization term therefore guides the motion toward well-conditioned postures while preserving the flexibility offered by kinematic redundancy.

\subsubsection{Terminal Cost}
The terminal cost drives the end-effector to converge to the desired goal pose at the end of the prediction horizon. Using the same Lie-algebraic pose error definition introduced in the pose tracking cost, the terminal pose error is denoted by $\boldsymbol{\xi}(H)\in\mathbb{R}^6$, and the terminal cost is defined as
\begin{equation}
    J_{\mathrm{term}}
    =
    \tfrac{1}{2}\,
    \boldsymbol{\xi}(H)^\top
    \boldsymbol{Q}_H\,
    \boldsymbol{\xi}(H),
\end{equation}
where $\boldsymbol{Q}_H \succ 0$ is a terminal weighting matrix, ensuring accurate convergence to the target pose at the final step.

\subsection{Trajectory Optimization Process}
With the cost components constructed in the previous subsection, the hard constraints in the original problem~\eqref{eq:opt_problem} are incorporated  into the objective through penalty terms. This reformulation converts the constrained optimal control problem into an unconstrained optimization over the control sequence $\mathbf{U}$, which can be efficiently solved using a SMPC framework. 

\begin{algorithm}[t]
  \caption{Sampling-based MPC}
  \label{alg:smpc}
  \KwIn{
    $M$: the number of samples,
    $H$: prediction horizon,
    $\mathcal{D}$: zero-mean sampling distribution
  }
  \KwOut{$\mathbf{U}^\star$: the optimal control sequence}
  \LinesNumbered

  Initialize nominal control sequence 
  $\boldsymbol{\bar{U}} = \mathbf{0}$\;

  \For{$m \gets 1$ \textbf{to} $M$}{
      Sample disturbances 
      $\{\boldsymbol{\epsilon}_k^{(m)}\}_{k=0}^{H-1} \sim \mathcal{D}$\;

      Construct sampled control sequence
      $\mathbf{U}^{(m)} = \boldsymbol{\bar{U}} + \boldsymbol{\epsilon}^{(m)}$\;

      Roll out dynamics~(\ref{eq:dynamic}) to obtain the sampled state trajectory
      $\{ \boldsymbol{q}_k^{(m)} \}_{k=0}^{H}$ together with 
      $\{\dot{\boldsymbol{q}}_k^{(m)}, \ddot{\boldsymbol{q}}_k^{(m)}\}$\;      

      Compute link and EE poses via forward kinematics~(\ref{eq:forward_kinematics})\;

      Evaluate the total trajectory cost $S^{(m)}$~(\ref{eq:total_trajectory_cost})\;
  }

  Compute minimum cost 
  $S_{\min} = \min_m S^{(m)}$\;

  \For{$m \gets 1$ \textbf{to} $M$}{
      Compute importance weight using~(\ref{eq:weighting}):
      \[
         w^{(m)} = \text{SoftWeight}(S^{(m)}, S_{\min}, \lambda)\;.
      \]
  }

  Update the optimal control sequence
  $\mathbf{U}^\star$ using~(\ref{eq:u_update})\;

  \Return $\mathbf{U}^\star$ 
\end{algorithm}

At each time step, the SMPC samples a batch of $M$ candidate control sequences around a nominal sequence. Each control sequence is expressed as
\begin{equation}
    \small
    \mathbf{U}^{(m)}
    =
    \left[\boldsymbol{u}_0^{(m)}, \boldsymbol{u}_1^{(m)}, \dots, \boldsymbol{u}_{H-1}^{(m)}\right],
    \qquad m = 1,\dots,M,
\end{equation}
where each control input is obtained by perturbing the nominal action:
\begin{equation}
  \boldsymbol{u}_k^{(m)} = \boldsymbol{\bar{u}}_k + \boldsymbol{\epsilon}_k^{(m)}, 
    \qquad
    \boldsymbol{\epsilon}_k^{(m)} \sim \mathcal{D},
\end{equation}
with $\boldsymbol{\bar{u}}_k$ the nominal action and $\mathcal{D}$ a zero-mean sampling distribution.

For each sampled sequence $\mathbf{U}^{(m)}$, the system dynamics are rolled out over the prediction horizon to generate a candidate state trajectory 
$\{\boldsymbol{q}_k^{(m)}\}_{k=0}^{H}$, while evaluating all running cost 
components along the way and the terminal cost at $k=H$. This yields the total 
trajectory cost
\begin{equation}
    S^{(m)}
    =
    \ell_f\!\left(\boldsymbol{T}_H^{(m)}, \boldsymbol{T}_g\right)
    \;+\;
    \sum_{k=0}^{H-1} \ell\!\left(\boldsymbol{q}_k^{(m)}, \boldsymbol{x}_k^{(m)}, \boldsymbol{u}_k^{(m)}\right). 
    \label{eq:total_trajectory_cost}
\end{equation}

After computing the trajectory cost $\{S^{(m)}\}_{m=1}^M$, a soft weighting 
scheme is applied to combine the candidates. Each trajectory is assigned an 
importance weight
\begin{equation}
  w^{(m)}
  =
  \frac{
      \exp\!\left[-\frac{1}{\lambda}\left(S^{(m)} - S_{\min}\right)\right]
  }{
      \sum_{j=1}^{M}
      \exp\!\left[-\frac{1}{\lambda}\left(S^{(j)} - S_{\min}\right)\right]
  },
  \label{eq:weighting}
\end{equation}
where $S_{\min} = \min_{j=1,\dots,M} S^{(j)}$ is the lowest cost in the batch, and $\lambda>0$ is a temperature parameter that determines how sharply the 
weights concentrate around low-cost trajectories.

Using the computed weights, the optimal control sequence is refined by taking a 
weighted average over the sampled disturbances:
\begin{equation}
  \mathbf{U}^\star
  =
  \sum_{m=1}^{M} w^{(m)} \mathbf{U}^{(m)}
  =
  \boldsymbol{\bar{U}}
  +
  \sum_{m=1}^{M} w^{(m)} \boldsymbol{\epsilon}^{(m)},
  \label{eq:u_update}
\end{equation}
where
\begin{equation}
  \boldsymbol{\bar{U}}
  =
  [\boldsymbol{\bar{u}}_0,\dots,\boldsymbol{\bar{u}}_{H-1}],
  \qquad
  \boldsymbol{\epsilon}^{(m)}
  =
  [\boldsymbol{\epsilon}_0^{(m)},\dots,\boldsymbol{\epsilon}_{H-1}^{(m)}].
\end{equation}

The trajectory optimization process is summarized in Algorithm~\ref{alg:smpc}. Following the receding-horizon principle of MPC, only the first control input $\boldsymbol{u}_0^\star$ of the optimized sequence is executed. Specifically,  $\boldsymbol{u}_0^\star$ is applied to the robot dynamics~(\ref{eq:dynamic}) to obtain the next joint state via one-step forward integration, and the resulting updated state is sent to the low-level controller. The entire sampling--rollout--weighting procedure is then repeated at the next time step using the new state feedback. It is worth noting that the sampling and rollout steps (Algorithm~\ref{alg:smpc}, lines~2--7) are mutually independent and can be implemented fully in parallel, making the proposed SMPC framework highly amenable to modern GPU acceleration.

\section{Results}
This section presents both quantitative benchmark comparisons and real-world experimental results. We first evaluate the proposed system in simulation, assessing the performance of the mapping module and the sampling-based motion planner. We then validate the full pipeline on a 7-DoF robotic manipulator (Flexiv Rizon~4) to demonstrate its effectiveness and robustness in real-world scenarios. All experiments in this section are conducted on a computer equipped with an Intel i9-10900K CPU (3.7\,GHz) and an NVIDIA RTX~3090 GPU.

\subsection{Benchmark Comparisons}

\subsubsection{Mapping Evaluation}
To evaluate the proposed environment mapping algorithm, we conduct quantitative 
experiments on two types of publicly available datasets:
(i) the Flat Dataset%
\footnote{\url{https://projects.asl.ethz.ch/datasets/panoptic-mapping}}, 
which provides depth images synchronized with ground-truth poses, and 
(ii) the Dynablox Dataset%
\footnote{\url{https://projects.asl.ethz.ch/datasets/dynablox}}, 
which contains point clouds from a 128-beam OS0 LiDAR together with TF-based sensor poses. These datasets correspond to the two sensing modalities supported by our framework, enabling us to assess mapping performance on both 
depth-camera and LiDAR-based inputs. We compare our method against the state-of-the-art \texttt{GIE-Mapping}~\cite{GIE-mapping} under identical experimental settings.

\begin{figure}[t]
  \centering
  \subfloat[]{\includegraphics[width=1.8in]{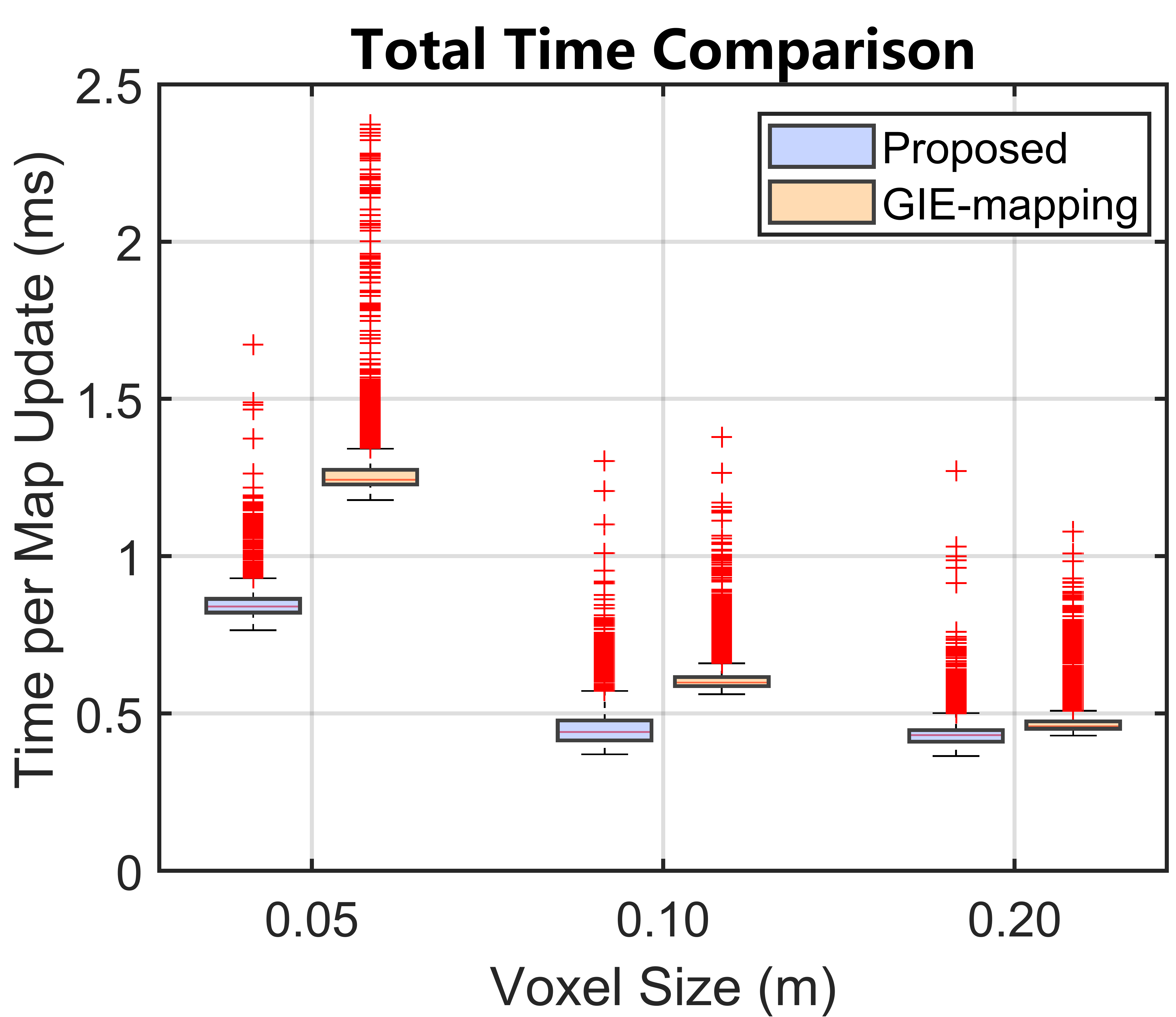}
  \label{fig_first_case}}
  \subfloat[]{\includegraphics[width=1.8in]{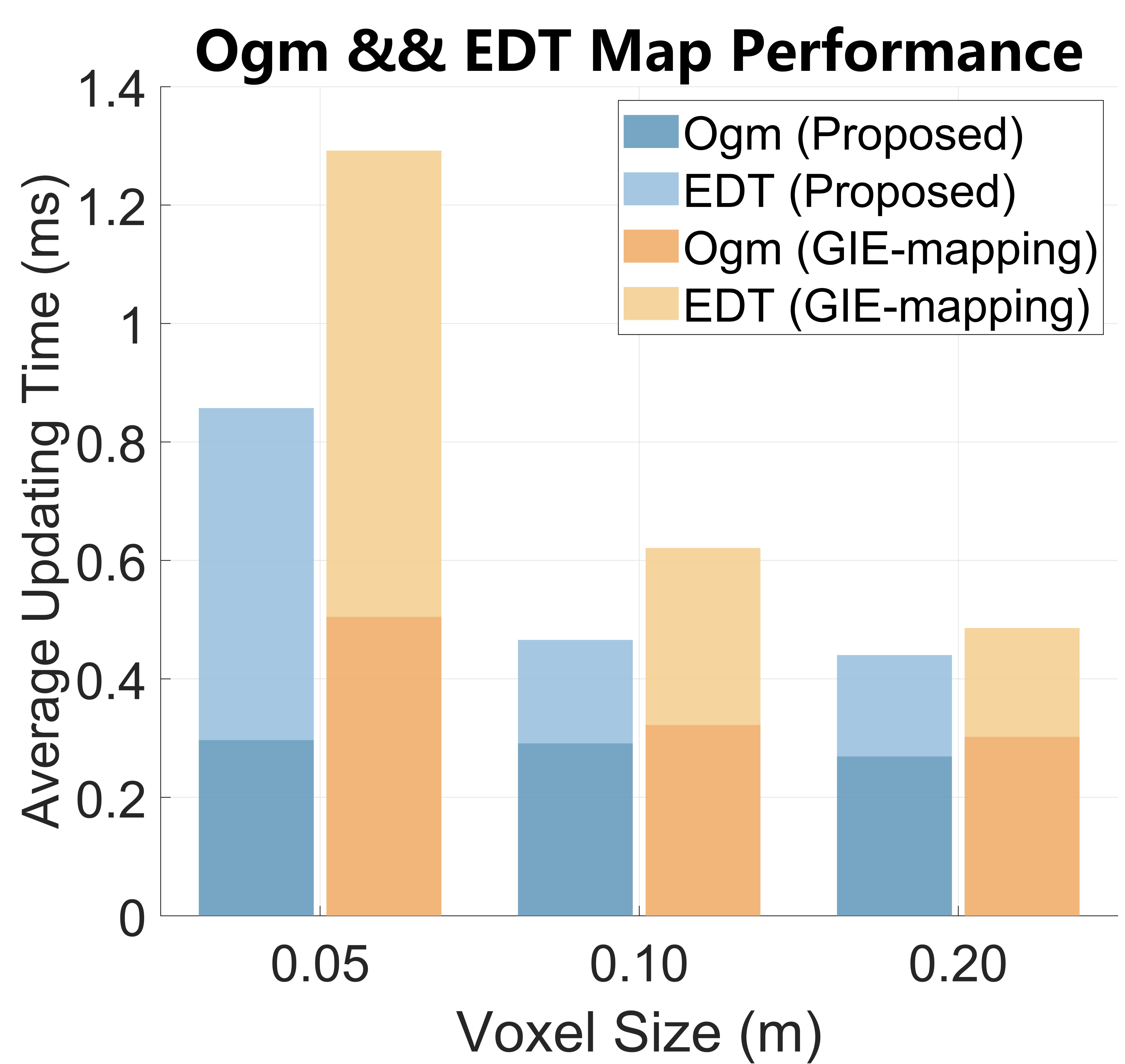}
  \label{fig_second_case}}
  \caption{Performance with respect to voxel size on the Flat Dataset. (a) Total mapping time, consisting of OGM and the EDT. (b) Average updating time of OGM and EDT. The EDT is computed within a fixed local volume of 
  $8\,\text{m} \times 8\,\text{m} \times 3\,\text{m}$ centered on the robot.
  } 
  \label{fig_map_performance_camera}
\end{figure}

For the Flat Dataset, the EDT is computed within a fixed local volume of 
$8\,\mathrm{m} \times 8\,\mathrm{m} \times 3\,\mathrm{m}$ centered on the robot. Fig.~\ref{fig_map_performance_camera}(a) reports the total mapping time (OGM + EDT) for voxel sizes of $0.05\,\mathrm{m}$, $0.10\,\mathrm{m}$, and $0.20\,\mathrm{m}$. Across all resolutions, the proposed method consistently achieves lower total update time than \texttt{GIE-Mapping}. As shown in Fig.~\ref{fig_map_performance_camera}(b), this improvement is primarily driven by a substantially faster EDT update, while the OGM stage also exhibits a slight reduction in computation time due to a streamlined system implementation.

\begin{figure}[t]
  \centering
  \subfloat[]{\includegraphics[width=1.8in]{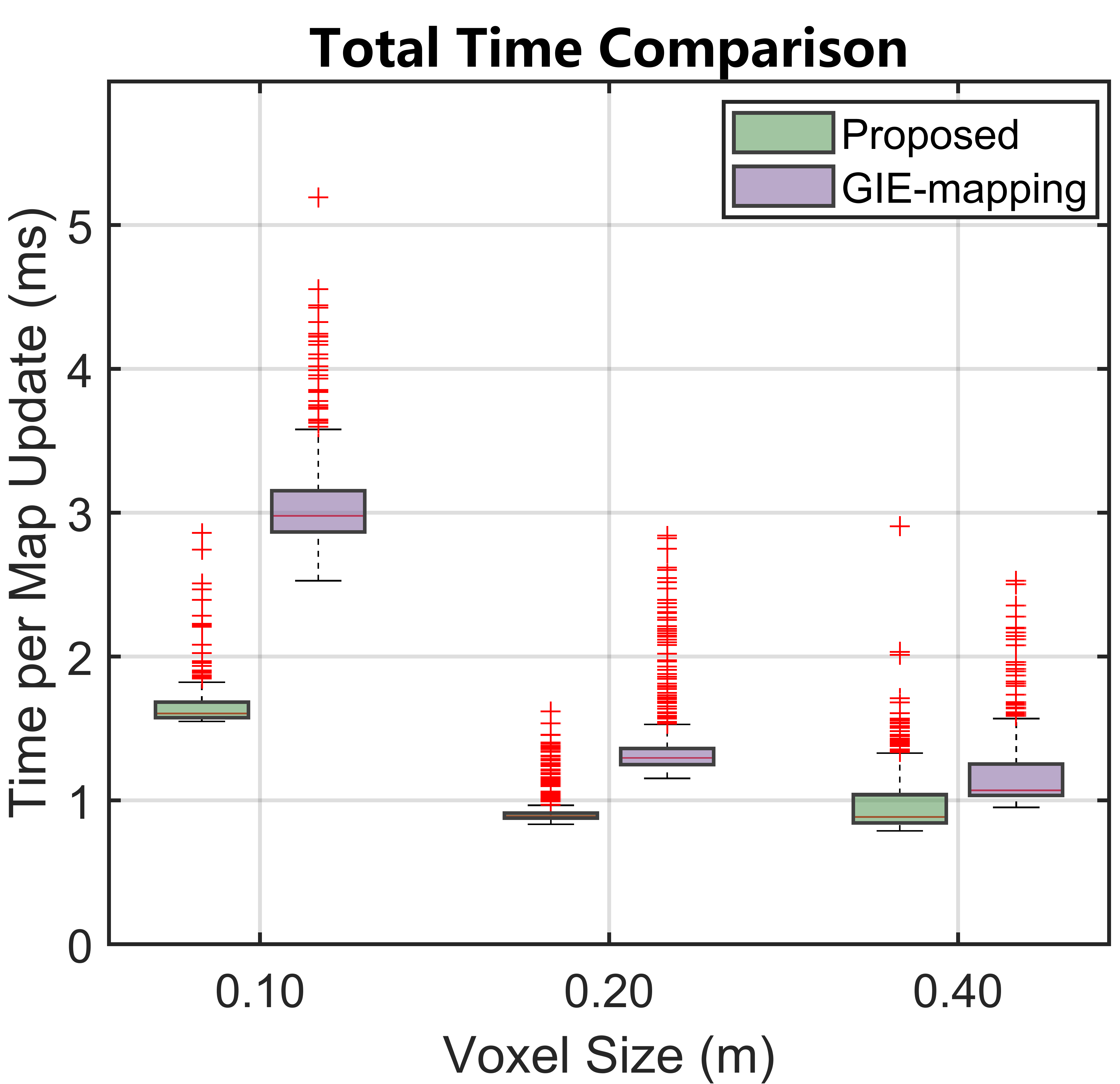}
  \label{fig_first_case}}
  \subfloat[]{\includegraphics[width=1.8in]{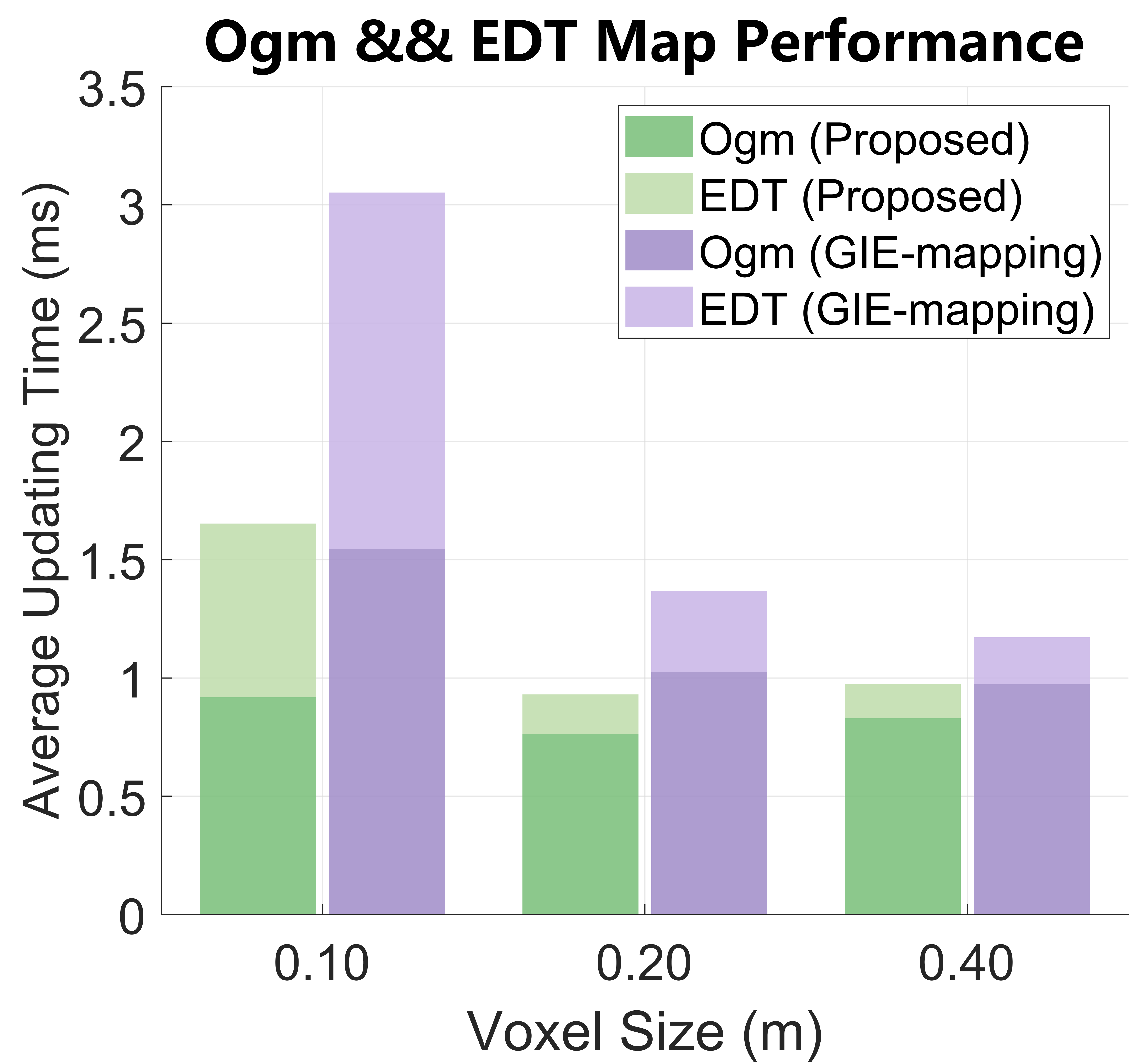}
  \label{fig_second_case}}
  \caption{Performance with respect to voxel size on the Dynablox Dataset. (a) Total mapping time (OGM + EDT). (b) Average updating time of OGM and EDT. The EDT is computed in a fixed $20 \times 20 \times 6\,\text{m}^3$ local volume.
  } 
  \label{fig_map_performance_lidar}
\end{figure}

For the Dynablox Dataset, a larger EDT volume of $20\,\mathrm{m} \times 20\,\mathrm{m} \times 6\,\mathrm{m}$ is used to accommodate the wide-range outdoor LiDAR environment. As shown in Fig.~\ref{fig_map_performance_lidar}(a), our method achieves lower total mapping time across voxel sizes of $0.10\,\mathrm{m}$, $0.20\,\mathrm{m}$, and $0.40\,\mathrm{m}$. Fig.~\ref{fig_map_performance_lidar}(b) further reveals that the performance gain arises from reduced computation in both OGM and EDT, with the EDT module contributing the dominant portion of the improvement. Notably, the proposed EDT update remains efficient even at fine voxel resolutions, where traditional EDT methods typically incur significant computational overhead. 

Overall, the results on both datasets demonstrate that the proposed framework is efficient and scalable across sensing modalities and voxel resolutions, achieving reductions in both OGM and EDT update time compared with \texttt{GIE-Mapping}. These efficiency gains are particularly important for robotic manipulation and motion-planning tasks operating in dynamic scenes, where high-resolution maps and fast update rates are essential for safe and reactive motion.

\subsubsection{Motion Planning Evaluation}
We evaluate the motion-planning performance of the proposed SMPC 
planner and compare it against two representative baseline methods: STORM \cite{Storm}, and RRTConnect from the OMPL library \cite{OMPL}, which is integrated within the MoveIt\footnote{\url{https://moveit.ai}} motion planning framework. Since neither STORM nor RRTConnect supports online environment mapping, all planners are evaluated under a known obstacle environment, allowing us to focus on a direct comparison of their motion-planning capabilities. For a fair comparison, the proposed method and STORM are evaluated using the same number of samples per replanning iteration ($M = 512$).

\begin{figure}[t]
  \centering
  \includegraphics[width=3.0in]{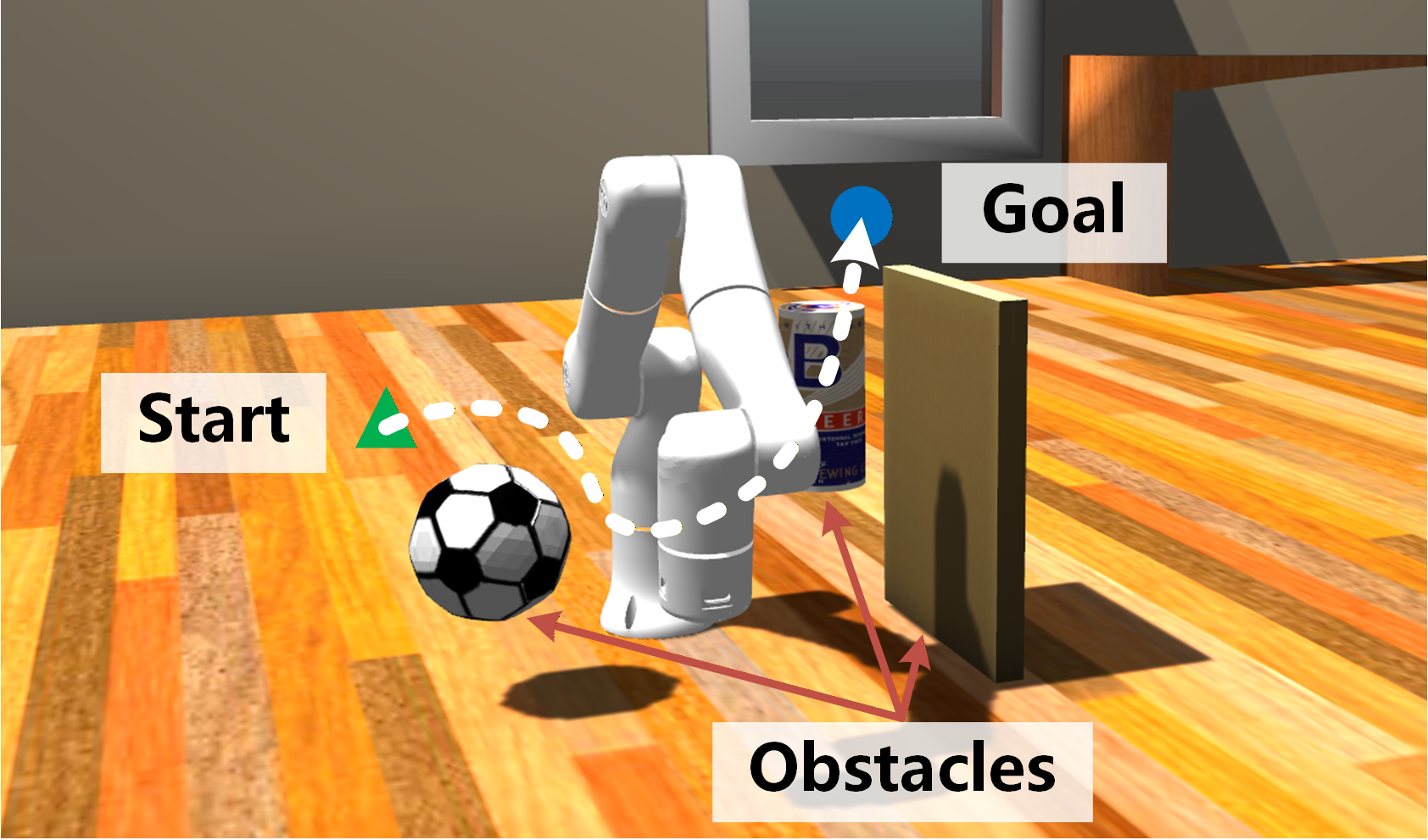}
  \caption{Benchmark motion-planning scenario in the Gazebo simulation.}
  \label{fig_sim_scene}
\end{figure}

The benchmark scenario is constructed in the Gazebo simulation platform, as illustrated in Fig.~\ref{fig_sim_scene}. In this scenario, the robot is required to plan a collision-free motion from a given start joint configuration to a target pose in the presence of multiple obstacles. To quantitatively assess performance, we evaluate each method using the following metrics: planning time, motion time, path length and final pose error (position error and orientation error).

\paragraph{Pose Error Metrics}
The pose error provides a direct measure of goal-reaching accuracy and is computed at the final converged solution. It is measured in terms of both position error and orientation error. The position error is defined as the Euclidean distance between the desired end-effector position $\mathbf{p}_{\text{g}}$ and the achieved position $\mathbf{p}_{\text{final}}$:
\begin{equation}
e_{\text{pos}} = \left\| \mathbf{p}_{\text{g}} - \mathbf{p}_{\text{final}} \right\|_2.
\end{equation}

Similarly, the orientation error is computed as the angular distance between the desired quaternion $\mathbf{q}_{\text{g}}$ and the achieved quaternion $\mathbf{q}_{\text{final}}$, and is defined as
\begin{equation}
e_{\text{ori}} = 2 \arccos \left( \left| \mathbf{q}_{\text{g}}^\top \mathbf{q}_{\text{final}} \right| \right),
\end{equation}
which represents the minimal rotation angle between the two orientations.

\paragraph{Convergence and Termination Criteria}
For the proposed method and STORM, which are both MPC-based planners, replanning is performed iteratively at a fixed frequency (50~Hz in our settings). A solution is considered converged when the following two conditions are simultaneously satisfied over multiple consecutive iterations:  
(i) the relative improvement of the objective value falls below a predefined threshold, and  
(ii) the position error is below a predefined tolerance (10~mm in our settings).

Once both conditions are satisfied for $N_{\text{stable}} = 5$ consecutive iterations, the planner is deemed to have converged and the planning process is terminated. The final path length is then computed as the accumulated joint-space motion, measured as the sum of angular displacements (in radians) across all joints, up to the convergence point. The accumulated trajectory execution time up to convergence is reported as the motion time. In contrast, RRTConnect is a single-shot sampling-based planner without iterative replanning. Therefore, for RRTConnect, the pose error, path length and motion time are directly computed from the resulting planned trajectory, without applying the above convergence criteria.

The experimental results are summarized in Table~\ref{tab:motion_planning_results}, with all metrics averaged over multiple runs to account for the stochastic nature of sampling-based planners. The proposed method achieves the shortest planning time among all baselines, which can be attributed to its fully parallelized, GPU-centric implementation of sampling, rollout generation, and cost evaluation. Although STORM also leverages GPU acceleration, its optimization pipeline is largely implemented at the Python level using PyTorch operators, which can introduce additional kernel launch and scheduling overhead. In addition to improved planning efficiency, the proposed method produces shorter joint-space path lengths compared to both STORM and RRTConnect, indicating more direct and smoother joint motions with reduced unnecessary joint excursions during execution. Furthermore, it achieves the shortest motion time and the smallest final pose error at convergence, reflecting faster execution toward the desired end-effector pose and more accurate goal reaching. These results highlight the effectiveness of the proposed Lie-algebra-based pose error formulation, which provides a geometrically consistent error metric on $\mathrm{SE}(3)$. By coupling translational and rotational errors in a unified geometric representation, the proposed formulation improves convergence speed and pose tracking accuracy within the SMPC framework.

\begin{figure}[]
  \centering
  \includegraphics[width=3.0in]{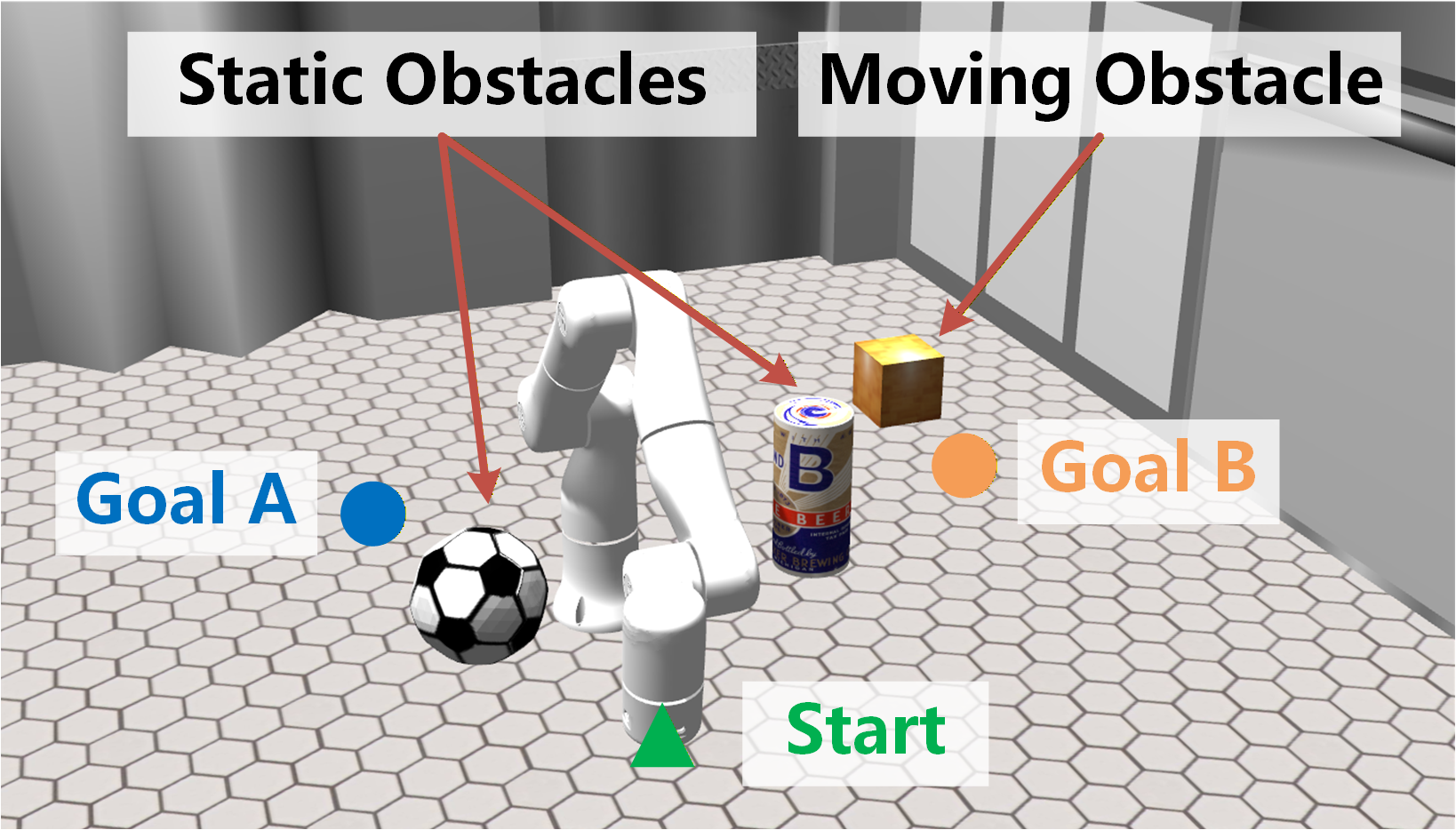}
  \caption{Integrated system evaluation scenario in the Gazebo simulation with static and dynamic obstacles.}
  \label{fig_sim_scene_dynamic}
\end{figure}

\begin{table}[t]
  \centering
  \caption{Quantitative comparison of motion-planning performance across different methods.}
  \label{tab:motion_planning_results}
  \begin{tabular}{lccc}
  \toprule
  \textbf{Metric} & \textbf{RRTConnect} & \textbf{STORM} & \textbf{Proposed} \\
  \midrule
  Planning Time (ms)      & 61.202  & 18.875 & \textbf{5.322} \\
  Motion Time (s)         & 13.720  & 16.580  & \textbf{12.412} \\
  Path Length (rad.)      & 13.153  & 15.316  & \textbf{12.341}  \\
  Position Error (mm)     & 3.674   & 9.299   & \textbf{2.778}  \\
  Orientation Error (rad.) & 0.104 & 0.327 & \textbf{0.049} \\
  \bottomrule
  \end{tabular}
\end{table}

\subsubsection{Integrated System Evaluation}
To further assess the proposed approach, we conduct an integrated system evaluation of the full pipeline that integrates online mapping and motion planning. The experiments are performed in a cluttered Gazebo simulation environment, as shown in Fig.~\ref{fig_sim_scene_dynamic}. In this scenario, the environment is initially fully unknown, and depth measurements are acquired online from a simulated depth camera sensor to construct the environment representation. The map is constructed using a voxel size of 0.02~m and maintains a $3 \times 3 \times 0.5~\mathrm{m}^3$ local EDT within a workspace centered around the robot. The motion planner runs at 50~Hz with $M = 512$ samples per iteration, continuously incorporating updated map information during execution.

\begin{figure*}[t]
  \centering
  \includegraphics[width=\textwidth]{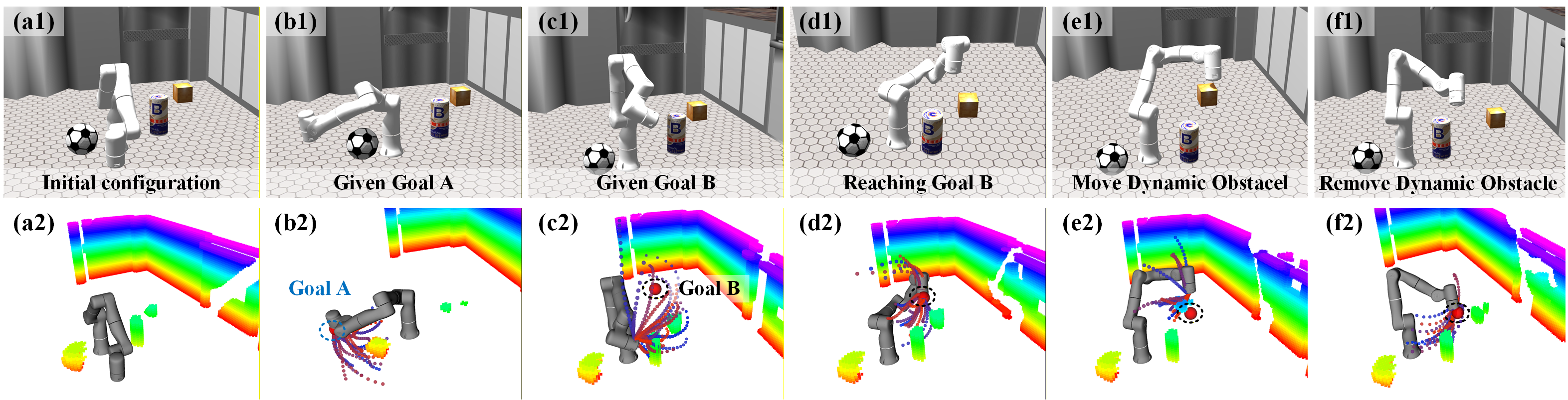}
  \caption{Sequential snapshots of the proposed integrated system during task execution in simulation. The top row (a1--f1) shows the robot execution and environment changes in Gazebo, while the bottom row (a2--f2) illustrates the corresponding online mapping and replanning behavior. From left to right, the robot starts from an initial configuration (a1, a2) and plans toward Goal~A (b1, b2), followed by a transition to Goal~B (c1--d1, c2--d2). After reaching Goal~B, a dynamic obstacle begins to move within the workspace, triggering real-time replanning and obstacle avoidance (e1, e2). Once the dynamic obstacle is removed, the robot resumes motion and returns to Goal~B, successfully completing the task (f1, f2). In the bottom row, the colored trajectories represent sampled motion rollouts generated by the SMPC planner.}
  \label{fig_sim_plan_process}
\end{figure*}

The task is designed as a multi-stage scenario. Starting from an initial configuration, the robot first plans and moves toward Goal~A, and subsequently transitions to Goal~B. A dynamic obstacle is present in the workspace throughout the task and is modeled as a cubic block of size $10\,\mathrm{cm} \times 10\,\mathrm{cm}$, whose motion is manually controlled via keyboard input. After the robot reaches Goal~B, the dynamic obstacle begins to move and interferes with stable convergence to the target pose. To ensure safety, the end-effector responds online by replanning and avoiding the moving obstacle in real time. Once the obstacle is moved away from the vicinity of the target region, the robot resumes motion and converges to Goal~B again. Representative snapshots of this process, including both robot execution and the corresponding online mapping and replanning behavior, are shown in Fig.~\ref{fig_sim_plan_process}. Thanks to the robot-masked distance field formulation, together with the reactive replanning capability of the SMPC planner, the robot is able to carry out the task reliably under online mapping and dynamic obstacle interactions.

We further report system-level runtime performance under the above mapping and planning configuration. The proposed SMPC planner achieves an average trajectory planning time of 5.653~ms per replanning iteration, while the average mapping update time is 0.892~ms per iteration. As a result, the combined mapping and planning pipeline can operate at update rates exceeding 150~Hz in this setting. These results demonstrate that the proposed framework can respond promptly and safely to dynamic obstacles during task execution.

\subsection{Real-world Experiments}
\begin{figure}[t]
  \centering
  \includegraphics[width=3.25in]{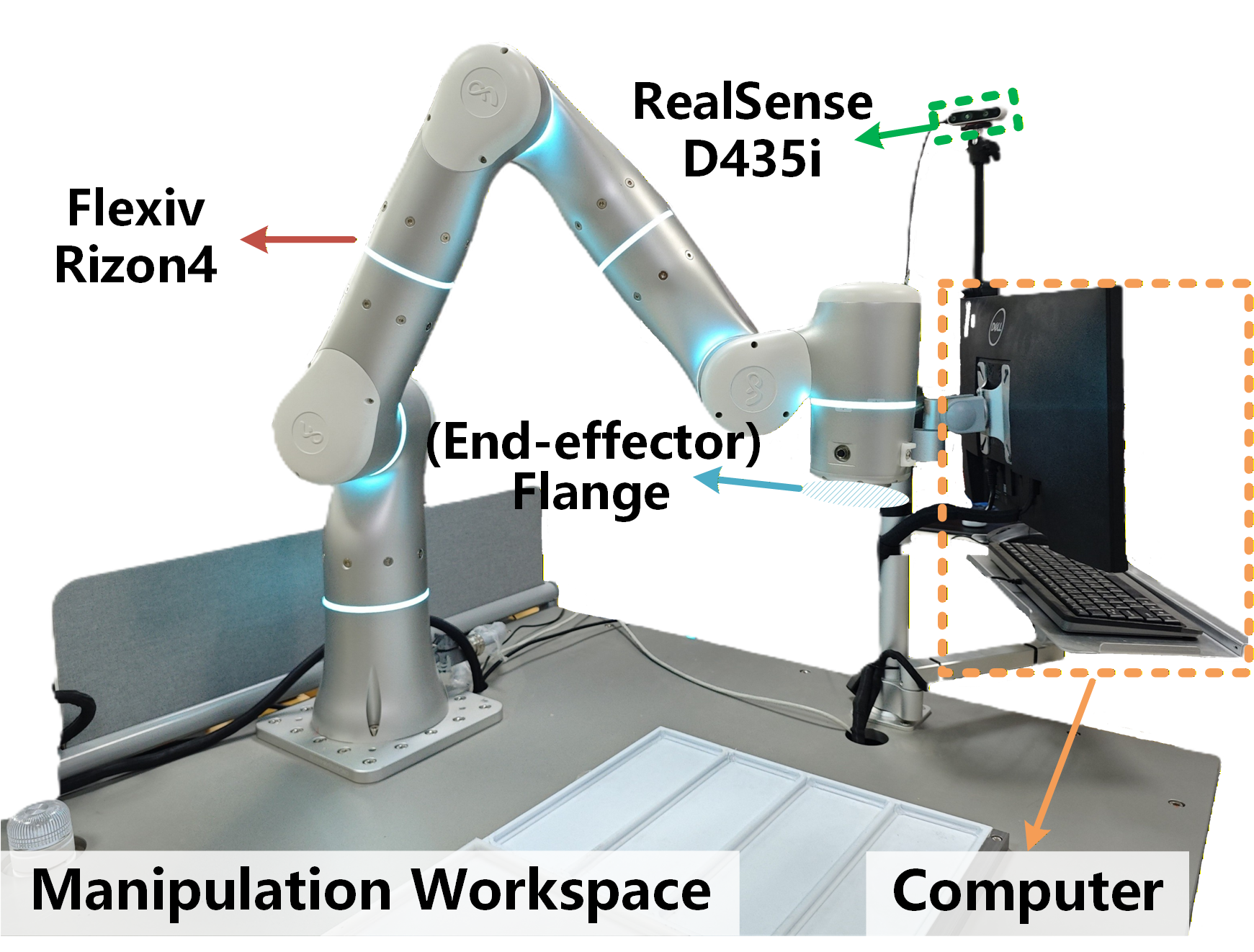}
  \caption{The 7-DoF robotic manipulator platform (Flexiv Rizon~4) used for real-world system evaluation.}
  \label{fig_real_platform}
\end{figure}

We further validate the proposed approach through real-world experiments on a physical robotic platform, focusing on the performance of the complete system under realistic operating conditions. The experiments are conducted on a 7-DoF Flexiv Rizon~4 robotic manipulator, equipped with an Intel RealSense D435i depth camera for online perception of the manipulation workspace, as shown in Fig.~\ref{fig_real_platform}. The camera is mounted in an eye-to-hand configuration and extrinsically calibrated with respect to the robot base-link frame. In our setup, the robot end-effector is defined at the flange frame. The online mapping and motion planning modules are configured identically to those used in the integrated system evaluation described in Sec.~V-A.3.

\begin{figure}[t]
  \centering
  \includegraphics[width=3.25in]{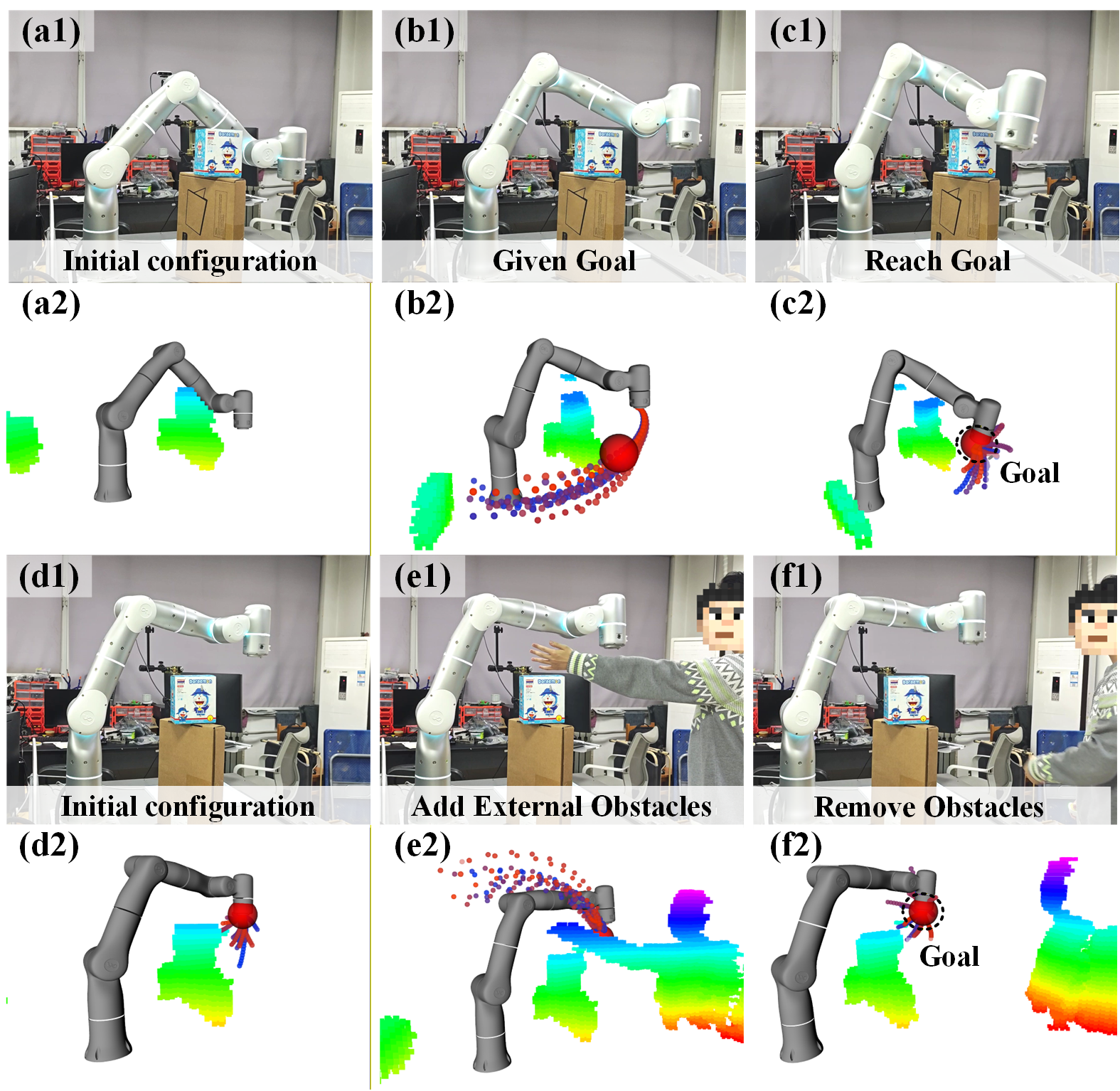}
  \caption{Snapshots of real-world experiments with the proposed system. (a)--(c): Online point-to-point motion planning in the presence of unknown obstacles. (d)--(f) Reactive motion planning with dynamic obstacles caused by a human arm moving within the camera field of view.}
  \label{fig_real_plan_process}
\end{figure}

As shown in Fig.~\ref{fig_real_plan_process}, two representative real-world scenarios are evaluated. In the first scenario (a)--(c), the task is formulated as point-to-point motion planning in a workspace with previously unknown obstacles, where the robot successfully reaches the target end-effector pose based on online perception. In the second scenario (d)--(f), dynamic obstacles are introduced by a human arm moving within the camera field of view. The robot reacts by updating the map and replanning its motion online to avoid collisions, and subsequently resumes motion toward the target pose after the human arm is removed.

Overall, the real-world experiments demonstrate that the proposed system operates reliably and safely in environments with both static and dynamic obstacles. The robot reactively adapts its motion in response to previously unseen obstacles and resumes goal-directed execution once the obstacles move away from the robot. These results validate the effectiveness of the proposed framework for integrated online perception, mapping, and motion planning in real-world scenarios.

\section{Conclusion}
In this paper, we present an integrated framework that combines online environment perception and sampling-based MPC to enable real-time, reactive motion generation for robotic manipulation in dynamic and previously unknown environments. A gather-then-transform Euclidean distance field construction is employed to reduce memory overhead and support low-latency distance queries by avoiding expensive tensor transpositions. Building on this representation, a robot-masked distance field update mechanism is incorporated to explicitly exclude the robot body from the environment map, preventing false self-collision detections while remaining well suited to sampling-based, gradient-free planning. Furthermore, a Lie-algebra-based pose error formulation on $\mathrm{SE}(3)$ provides a geometrically consistent optimization objective, facilitating efficient convergence and accurate goal reaching. Extensive simulation and real-world experiments demonstrate the practicality of the proposed framework and its ability to operate reliably and responsively in cluttered, initially unknown environments.

\bibliographystyle{Bibliography/IEEEtranTIE}
\bibliography{Bibliography/IEEEexample}\ 

\end{document}